\documentclass[10pt,journal,letterpaper,compsoc]{IEEEtran}

\ifCLASSOPTIONcompsoc
  \usepackage[nocompress]{cite}
\else
  \usepackage{cite}
\fi

\ifCLASSINFOpdf
     \usepackage[pdftex]{graphicx}
     \graphicspath{{./png/}}
     \DeclareGraphicsExtensions{.pdf,.png}
\else
    \usepackage[dvips]{graphicx}
    \DeclareGraphicsExtensions{.eps}
\fi

\usepackage[cmex10]{amsmath}

\usepackage{array}

\usepackage{fixltx2e}

\usepackage{url}

\usepackage[table]{xcolor}
\definecolor{Gray}{gray}{0.85}
\usepackage{color}

\newcommand{\cellem}{\cellcolor{gray}\color{white}}

\usepackage[ulem=normalem]{changes}

\usepackage{multirow}
\usepackage{footnote}
\usepackage[singlelinecheck=false]{caption}
\usepackage{booktabs}

\begin{document}
\title{Dense Correspondences Across Scenes and Scales}

\author{Moria Tau and Tal~Hassner
\IEEEcompsocitemizethanks{\IEEEcompsocthanksitem  M. Tau is with the Department of Mathematic and Computer Science, The Open University of Israel, Israel.%
\IEEEcompsocthanksitem T. Hassner is with the Department
of Mathematic and Computer Science, The Open University of Israel, Israel.\protect\\
E-mail: hassner@openu.ac.il}
\thanks{}}

\markboth{}
{Hassner and Tau: Dense Correspondences Across Scenes and Scales}

\IEEEcompsoctitleabstractindextext{%
\begin{abstract}
We seek a practical method for establishing dense correspondences between two images with similar content, but possibly different 3D scenes. One of the challenges in designing such a system is the local scale differences of objects appearing in the two images. Previous methods often considered only small subsets of image pixels;  matching only pixels for which stable scales may be reliably estimated. More recently, others have considered dense correspondences, but with substantial costs associated with generating, storing and matching scale invariant descriptors. Our work here is motivated by the observation that pixels in the image have contexts -- the pixels around them -- which may be exploited in order to estimate local scales reliably and repeatably. Specifically, we make the following contributions. (i) We show that scales estimated in sparse interest points may be propagated to neighboring pixels where this information cannot be reliably determined. Doing so allows scale invariant descriptors to be extracted anywhere in the image, not just in detected interest points. (ii) We present three different means for propagating this information: using only the scales at detected interest points, using the underlying image information to guide the propagation of this information across each image, separately, and using both images simultaneously. Finally, (iii), we provide extensive results, both qualitative and quantitative, demonstrating that accurate dense correspondences can be obtained even between very different images, with little computational costs beyond those required by existing methods.
\end{abstract}
}

\maketitle

\IEEEdisplaynotcompsoctitleabstractindextext
\IEEEpeerreviewmaketitle

\section{Introduction}
Establishing correspondences between pixels in two images is a fundamental step in many computer vision applications. Typically, this is performed by either matching a sparse set of pixels, selected by a repeatable detection method (e.g., the Harris-Laplace~\cite{mikolajczyk2004scale}), or by matching all pixels in both images. Here we focus on the latter case, seeking a practical means for establishing dense correspondences across images of different scenes in different local scales.

\begin{figure}
\begin{center}
\includegraphics[width=0.9\linewidth,clip,trim = 0mm 0mm 60mm 0mm]{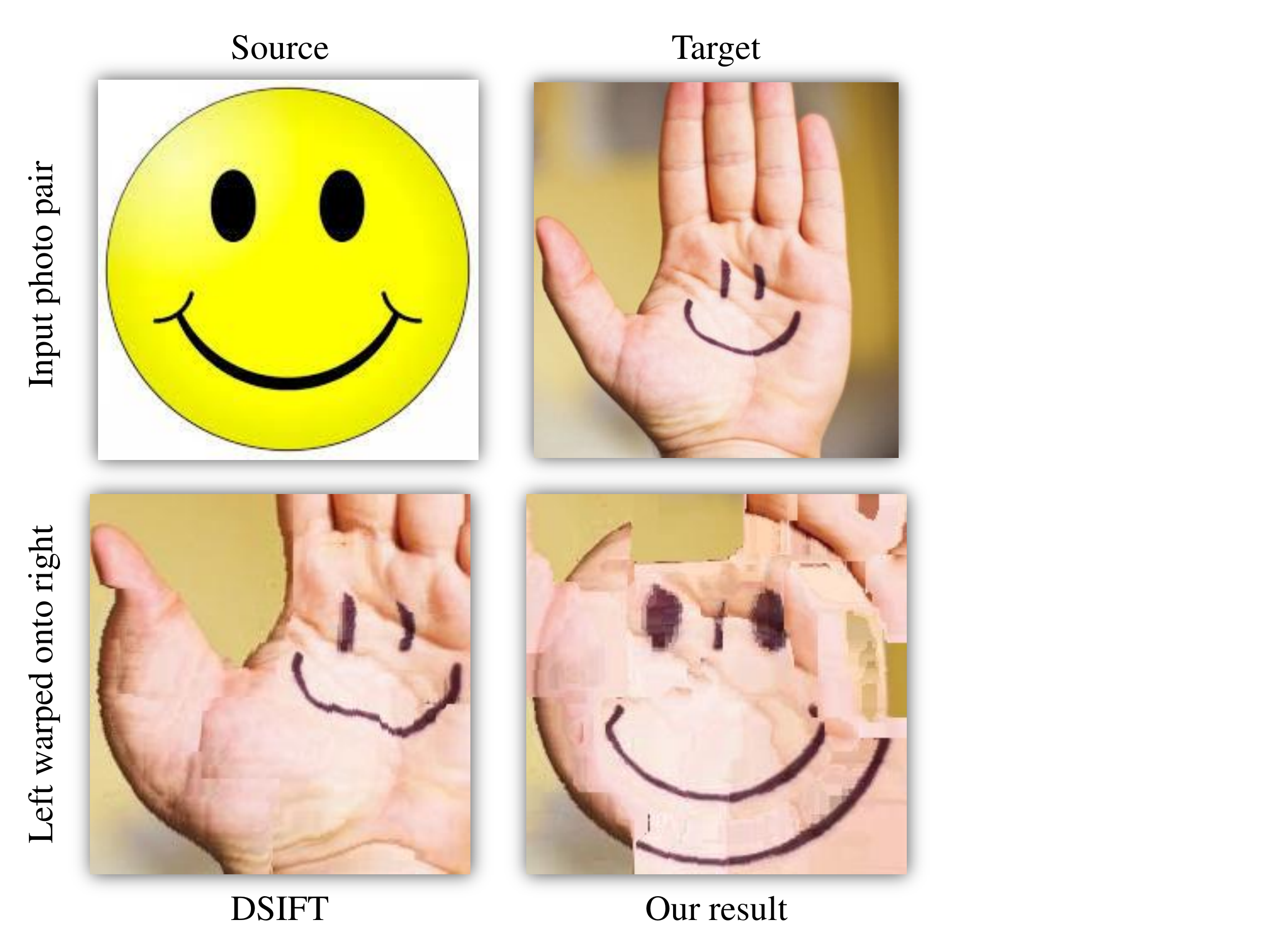}%
\caption{{\bf Dense correspondences between the same semantic content (``smiley'') in different scenes and different scales.} Top: Input images. Bottom: Results visualized by warping the colors of the ``Target'' photo onto the ``Source'' using the estimated correspondences from Source to Target. A good result has the colors of the Target photo, located in the same position as their matching semantic regions in the Source. Results show the output of the original SIFT-Flow method, using DSIFT without local scale selections (bottom left), and our method (bottom right).}\label{fig:teaser}
\end{center}
\end{figure}

Corresponding pixels are expected to reflect the same visual information. This information, however, may appear at different visual scales in different regions of each image: A car may be close to the camera in one photo, and far away in another; appearing large in the first and small in the second. All the while, buildings in the background remain at the same distance from the camera, appearing the same in both images. Sparse correspondence estimation methods seek stable scales, which can be repeatably detected in different images of the same scene, and which would allow extracting the same visual information regardless of the scales of the objects in the images. This approach, however, is only known to work well for very few pixels -- those where stable scales can be reliably detected~\cite{SIFT,mikolajczyk2002detection}.

Take, for example, the images in Fig.~\ref{fig:teaser} (Top). These present the same semantic content (a ``smiley''), appearing in very different scenes and in different scales. Densely matching the pixels in these two images is a problem made especially challenging due to the wide expanses of homogeneous regions, where stable scales are difficult to determine. In order to estimate correspondences, existing methods therefore make assumptions on the nature of the scenes, the photos, and the desired correspondences themselves.

Stereoscopic systems, for example, generally assume that the images being matched are of the same 3D scene, present objects in mostly the same scales, and were obtained under similar viewing conditions~\cite{bruhn2005lucas}. Recently, the same-scene assumption has been relaxed by the SIFT-Flow method of~\cite{liu2008sift,liu2011sift}. Although an important step, SIFT-Flow relies on the Dense-SIFT (DSIFT) descriptor of~\cite{vedaldi2010vlfeat}, and therefore implicitly assumes that visual information in both images appears at the same (arbitrarily selected) scale. More importantly, this scale assumption is the same for all pixels in both images; in essence, assuming a single global scale for the two images and so greatly limiting its applicability.

In the past few years, a number of methods have proposed to eliminate this same-scale assumption, thereby allowing for dense correspondences to be obtained under very general settings. These, however, are either designed to match images from the same scenes~\cite{kokkinos2008scale}, or require significant computation and storage in order to deal with unknown variations in scale~\cite{hassner2012sifts,qiu2014scalespasesift}.

In this paper we show that dense correspondences can be established reliably, even in challenging settings, such as those exemplified in Fig.~\ref{fig:teaser}, with little more computational and storage requirements than needed for the original SIFT-Flow algorithm.

Our work follows the observation that previous attempts to produce robust, dense descriptors, did so by treating each pixel {\em independently}, without considering the scales of other pixels in the image. We turn to those few pixels where scales have been reliably estimated, and use them in order to estimate scales for all other pixels. Realizing this idea, however, requires that we answer an important question: How should scales be propagated, from the few pixels where they were reliably determined to all others, in a way which would ensure repeatable scale assignments, and consequent accurate dense correspondence estimation, regardless of local scale changes? 

We answer this question by examining three means of propagating scale information across images, from detected key-points where scale is available to pixels where scales are not. Each of these methods considers progressively more information in order to more reliably propagate scales:
\begin{enumerate}
\item {\bf Geometric.} We propagated scale information from detected interest points by considering only the spatial locations where scales were detected (Sec.~\ref{sec:geometric}).
\item {\bf Image-aware.} Scales are propagated as above, but using image intensities in order to guide scale propagation. This is described in Sec.~\ref{sec:image}.
\item {\bf Match-aware.} Finally, in Sec.~\ref{sec:match} we consider the two images between which correspondences are to be estimated, propagating only the scales of pixels that were selected as (sparse) key-points in {\em both} images.
\end{enumerate}

\noindent We test these three approaches on a wide variety of qualitative and quantitative experiments, comparing them to the state-of-the-art. Our results show that more contextual information results in better correspondences. More importantly, they demonstrate our proposed approach to not only outperform existing methods, but to do so as efficiently as the original SIFT-Flow.

\section{Previous work}\label{sec:prev}

\noindent {\bf Why dense-flow?} Matching all the pixels of two images is a basic step in stereoscopic vision, and as such has been the subject of immense research from the early years of computer vision. Surveying the work on stereo correspondences is outside the scope of this paper, and we refer the reader to popular computer vision textbooks for descriptions of previous related work. A comprehensive treatment of this subject is provided in particular in~\cite{Hartley2004}.

In recent years, a new thread of work has sought to look beyond the single-scene settings of stereo systems, attempting to provide dense correspondences between images, even if they only share the same semantic content. Here, the motivation rose from the realization that by densely linking the pixels of two images, local, per-pixel information can be transferred from one image to the other. This information can then be used for a wide range of computer vision applications, including single-view depth estimation~\cite{hassner2006example,karsch2012depth,hassner2013single}, semantic labels and segmentation~\cite{liu2011nonparametric,rubinstein2013unsupervised}, image labeling and similarity~\cite{rubinstein2012annotation}, and even new-view synthesis~\cite{hassner2013viewing}.

In all cases described above, however, the same scale was assumed for the images involved. This, either by enforcing global alignment of the images (e.g.,~\cite{hassner2013viewing}) or by assuming that a large enough collection of images exists such that at least one will portray the same information in the same scales~\cite{rubinstein2013unsupervised}. The method presented here makes neither of these assumptions.\\

\noindent {\bf Scale-selection.} 
Objects appear in different scales in different images. Determining the correct scale at which an image portion must be processed has therefore been a long standing challenge in computer vision. Here we only briefly survey the vast literature on this subject, and we refer to~\cite{tuytelaars2008local,aanaes2012interesting} for more detailed discussions.

In his pioneering work, Lindeberg~\cite{lindeberg1998feature,lindeberg1999principles} was one of the first to suggest seeking image pixels which have well-defined, characteristic scales. He proposed using the Laplacian of Gaussian (LoG) function computed over image scales, which is covariant with the scale changes of the visual information in the image, and so allows extracting scale invariant descriptors.

In a subsequent work, Lowe~\cite{SIFT} proposed replacing the computationally expensive LoG function, with its Difference of Gaussians (DoG) approximation, in what has since become one of the standard de facto techniques for scale selection. Specifically, an image is processed by producing a 3D structure of $x,y$ and $scale$, using three sets of sub-octave, DoG filters. This structure is scanned in search of pixels with higher or lower values than their 26 space-scale neighbors (3$\times$3 neighborhood in the current scale and its two adjacent scales). The scale which provides these local extrema is selected as the characteristic scale for the pixel. 

These feature detectors, as well as others, select pixels as keypoints if such a characteristic scale can be selected. Some perform scale selection along with elimination of low-contrast pixels to obtain more reliable detections. One popular example is the Harris-Laplace detector~\cite{mikolajczyk2004scale}, which uses a scale-adapted Harris corner detector for spatial point localization and LoG filter extrema for scale selection. The detector performs these two steps iteratively, searching peaks both in space and in scale, and rejecting pixels with responses lower than a given threshold.\\

\noindent {\bf Dense-flow with changing scales.} A well known limitation of scale selection techniques is that they typically find reliable scales in only very few image pixels. In~\cite{mikolajczyk2002detection}, Mikolajczyk estimated that for a scale change factor of 4.4, as few as 38$\%$ of the pixels would be selected by a DoG scale selection criteria, of which only about 10.6$\%$ were actually correct. A bit later, Lowe, in~\cite{SIFT} estimated that only around 1$\%$ of an image's pixels provide stable features which allow for descriptor extraction and matching. If our goal is to obtain dense correspondences between two images, the obvious question becomes: how should scales be selected for the remaining overwhelming majority of the pixels in the two images?

In recent years there have been several solutions proposed to this problem. In~\cite{kokkinos2008scale}, image intensities around each pixel were transformed to log-polar coordinate systems. Doing so converted scale and rotation to translation. Translation invariance was then introduced by applying FFT, thus obtaining the Scale Invariant Descriptors (SID). Though SID descriptors were shown to be scale and rotation invariant, even on a dense grid, their use of image intensities directly implies that they are not well suited for matching images of different scenes~\cite{hassner2012sifts}.

The SIFT-Flow method of~\cite{liu2008sift,liu2011sift} provided a means for dense correspondence estimation on a dense grid. They represented pixels in the image using Dense-SIFT (DSIFT) descriptors~\cite{vedaldi2010vlfeat}, produced at a constant, manually selected scale. This provides some scale invariance -- due to the inherent robustness of the SIFT descriptors -- but does not address anything beyond small scale changes. 

In~\cite{hassner2012sifts}, the DSIFT descriptors used by the SIFT-Flow were replaced by the Scale-Less SIFT (SLS) representation. These are produced by first extracting at each pixel multiple SIFT descriptors, at multiple scales. The set of SIFTs extracted at a particular pixel was used to fit a linear subspace, represented using the subspace-to-point mapping of~\cite{BHZM:CVPR07:ANSS,basri2009general,basri2010approximate}. The SLS descriptors were shown to be highly robust to scale changes as well as allowing matching between different scenes, but the cost of this was a quadratic inflation in the descriptor size, making them difficult to apply in practice.

\begin{figure*}[!t]
\begin{center}
\includegraphics[width=\linewidth,clip,trim = 0mm 0mm 0mm 0mm]{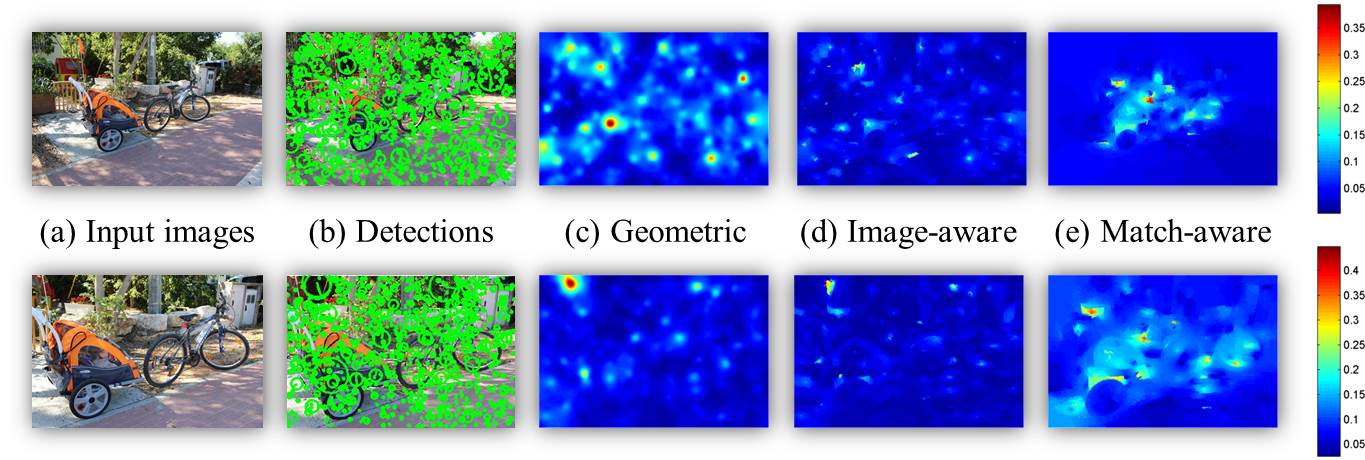}%
\caption{{\bf Visualizing three means of propagating scales.} (a) Two input images. (b) Sparse interest point detections, using the SIFT, DoG-based feature detector implemented by vlfeat~\cite{vedaldi2010vlfeat}. Each detection is visualized according to its estimated scale. (c-e) Per-pixel scale estimates, $S_I({\mathbf p})$, color-coded. (c) Geometric scale propagation (Sec.~\ref{sec:geometric}); (d) Image-aware propagation (Sec.~\ref{sec:image}); finally, (e), match-aware propagation, described in Sec.~\ref{sec:match}. Note how in (e) similar scale distributions are apparent for both images. On the right, color-bars provide legends for the actual scale values.}\label{fig:methods}
\end{center}
\end{figure*}

A different approach, somewhat related to our own work here, was taken by~\cite{trulls2013dense}. They too use SIFT-Flow as the matching engine, and either DSIFT or SID as the underlying representations. In their work, soft segmentation is first performed on images to be matched. When extracting descriptors, pixels contribute to the value of the descriptor in a manner which is inversely proportional to the likelihood of their belonging to the same segment as the keypoint for which the descriptor is produced. Thus, information from the background, or from other scales, has a limited effect on the values of the descriptor. This process requires that all descriptors are produced at the same scale, relying here on the segmentation to introduce scale-dependent information. Scales larger than the one used to extract the descriptors may therefor not be effectively represented. More importantly, it is unclear how segmentation affects scale, and vice versa, and so the limitations of this approach are not clear.

Rather than modify representations, Qiu et al. recently proposed a modified dense-flow estimation procedure~\cite{qiu2014scalespasesift}. Building on the cost function optimized by SIFT-Flow, they add terms reflecting the smoothness of scales. Specifically, they add a requirement that the relative scale of two neighboring pixels will be the same between their matching pixels in the other image. Though faster than both SID and SLS, their optimization is slower than the original SIFT-Flow. Moreover, their method does not allow computing scale invariant representations a priori, a desirable property when preprocessing is allowed or descriptors are used for applications other than dense correspondence estimation. Here we make a similar smooth scale assumption, yet employ it in preprocessing, rather than when estimating dense correspondences.

The method described here uses the original SIFT descriptors, varying the scales at which a descriptor is extracted in each pixel. It thus allows for correspondence estimation in the same computation and storage costs as the original SIFT-Flow as well as provides scale-invariant descriptors on a dense grid, usable beyond dense correspondence estimation applications.

\section{Propagating scales}\label{sec:propagation}
Scale-invariant correspondences (dense or otherwise) are typically achieved through scale selection. To establish {\em dense} correspondences, here, we seek {\em dense scale selection}: selecting scales for all the pixels in the image. 

Formally, the scale space of image $I(x,y)$, denoted by $L(x,y,\sigma)$, is defined by a convolution of $I(x,y)$ with a variable-scale Gaussian $G(x,y,\sigma)$~\cite{lindeberg1994scale}, where
\begin{equation}
L(x,y,\sigma)=G(x,y,\sigma)\star I(x,y)
\end{equation}
and
\begin{equation}
G(x,y,\sigma)=\frac{1}{2\pi\sigma^2}e^{-(x^2+y^2)/2\sigma^2}.
\end{equation}

\noindent The scale space of an image is scanned by multi-scale feature detectors, which seek space-scale locations $x,y,\sigma$ where stable scales can be determined reliably, typically by seeking extrema in a scale-selection function defined over $L(x,y,\sigma)$.

Most pixel coordinates, however, do not have such extreme values, and are therefore left without scale selection. In the texture rich images of Fig.~\ref{fig:methods}(a), for example, less than 0.1$\%$ of the pixels in each image were selected by the SIFT, DoG-based, feature detector, and assigned with scales (Fig.~\ref{fig:methods}(b)). Our goal is to use these few detected pixels and their scale assignments in order to estimate scales for all the remaining image pixels.

We define the {\em scale-map} $S_I(\mathbf{p})$, for pixel $\mathbf{p}=(x,y)$, of image $I$ as providing the scale $\sigma_{\mathbf p}$ associated with pixel coordinates $\mathbf{p}$ in $I$. Our goal can be stated as assigning scale values to all pixels in $S_I$. To this end, our key underlying assumption is stated as follows:\\

\noindent {\bf Assumption 0:} Similar pixels should have similar scales. \\

\noindent This assumption, of course, leaves the notion of similarity open for interpretation, as well as the means of assigning scales in practice. Formally, we express this general assumption by defining a global cost for a scale assignment, as follows:

\begin{equation}
C(S_I) = \sum_{\mathbf{p}}{\left(S_I(\mathbf{p})-\sum_{\mathbf{q}\in N(\mathbf{p})}(w_{\mathbf{p}\mathbf{q}}S_I(\mathbf{q}))\right)^2}.\label{eq:cost}
\end{equation}

\noindent Similar expressions have previously been proposed for image processing tasks ranging from segmentation (e.g.~\cite{shi2000normalized,weiss1999segmentation}, and others) to colorization~\cite{levin2004colorization} and depth estimation~\cite{guttmann2009semi}. Here, we assign scales to all image pixels by minimizing Eq.~\ref{eq:cost}, subject to the constraints expressed by the known scales -- the few pixels selected by a multi-scale feature detector, their positions in the image, and their assigned scales.

Intuitively, this cost interprets our assumption by requiring that the scale assigned to pixel $\mathbf{p}$ should be as similar as possible to a weighted average of the scales of its relevant similar pixels, denoted by $\mathbf{q}\in N(\mathbf{p})$. The weight $w_{\mathbf{p}\mathbf{q}}$, associated with each of these pixels $\mathbf{p}$ and $\mathbf{q}$, is often referred to as an {\em affinity function} and takes values which sum to one for all pixels $\mathbf{q}$. It reflects the degree to which the scale of one pixel is assumed to influence another. In the next sections we consider two alternatives for this function, based on different interpretations of pixel similarity.

\subsection{Geometric scale propagation}\label{sec:geometric}
Assuming that the only information available to us are the pixel locations and scales returned by a feature detector, we make the following ``geometric'' assumption, where pixel scales are influenced by the scales of their spatially neighboring pixels:\\

\noindent {\bf Assumption 1, Influence of feature geometry on scales:} Neighboring pixels (pixels with adjacent coordinates) should be assigned with the same scales. \\

\noindent This assumption can be interpreted as using a constant value for all affinity functions, or $w_{\mathbf{p}\mathbf{q}}=1/|N|$ ($|N|$ the number of spatial neighbors for each pixel). Our cost function is quadratic and our constraints are linear. This implies large, sparse systems of equations which may be solved using a range of existing solvers~\cite{shi2000normalized,levin2004colorization,guttmann2009semi}. 

Fig.~\ref{fig:methods}(c) presents the scale-maps produced for each image using geometric scale propagation. As scale assignments are covariant with the underlying scale changes in the images, their ranges are somewhat different. Scale color codes are therefore provided on the right of Fig.~\ref{fig:methods}. Visually, these maps may appear too noisy to be meaningful. In practice, as we show in Sec.~\ref{sec:results}, scales computed this way can still be beneficial for correspondence estimation.

\subsection{Image-aware scale propagation}\label{sec:image}
The use of constant affinity values is convenient whenever recomputing them for each image pair is impractical. Propagating scales using only the geometry of the feature point detections, however, ignores image intensities as valuable cues for scale assignment. We now consider the influence of intensities by revising our previous assumption.\\

\noindent {\bf Assumption 2, Influence of intensities on scales:} Neighboring pixels with similar intensities, should be assigned with similar scales. \\

\noindent This assumption can be expressed by assigning affinity values using the normalized cross-correlation of the intensities of the two pixels, or:

\begin{equation}
w_{\mathbf{p}\mathbf{q}} = 1+\frac{1}{\sigma_{\mathbf{p}^2}}\left((I(\mathbf{p})-\mu_{\mathbf{p}})(I(\mathbf{q})-\mu_{\mathbf{p}})\right).
\end{equation}

\noindent Here, $\mu_{\mathbf{p}}$ and $\sigma_{\mathbf{p}}$ are the mean and variance of the intensities in the neighborhood of pixel $\mathbf{p}$.

This expression has successfully been used in the past for image colorization in~\cite{levin2004colorization}. Earlier, it was shown to reflect a linearity assumption on the relation of color and intensities in~\cite{zomet2002multi} and~\cite{torralba2003properties}. By using it here, we assume a linear relation between intensities and {\em scales}, rather than color. That is, that $S_I(\mathbf{p})=a_{\mathbf{p}} I(\mathbf{p})+b_{\mathbf{p}}$ with the coefficients $a_{\mathbf{p}}$ and $b_{\mathbf{p}}$ being the same for all the pixels in the immediate neighborhood of $\mathbf{p}$. 

Of course, this is a simplifying assumption; the relation between scales and intensities can be more complex than a linear one. For our purposes, however, this assumption need not be strictly true: We only need for similar scales to be selected by considering similar intensity values in corresponding image regions. 

Fig.~\ref{fig:methods}(d) visualizes the scale-maps produced by image-aware propagation. These capture more of the underlying image appearance than the ones produced by the simpler geometry based method. In particular, the distribution of scale assignments for the two images has more regions in common, suggesting better repeatability. Still, quite a lot of both images includes non-matching scale assignments, which we attempt to minimize next.

\subsection{Match-aware scale propagation}\label{sec:match}
As evident in Fig.~\ref{fig:methods}(b), the sets of feature point detections in the two images are not identical. In fact, we expect only a small number of features to be correctly detected and common to both images (as discussed in Sec.~\ref{sec:prev}). Here, these few corresponding pixels are used to seed the scale-map assignment process: \\

\noindent{\bf Assumption 3, Influence of matching feature points:} When two images are being matched, scales should be assigned by considering feature point detections common to both images.\\

\noindent Rather than using all the detected feature points to seed the scale assignments, we first seek correspondences between the scale invariant descriptors, extracted at these sparse locations. This, in the same way that such correspondences are computed and used for parametric image alignment~\cite{SIFT}. We take the $20\%$ of the correspondences with the best closest to second-closest SIFT match ratio~\cite{SIFT}, and use only their scales to seed scale propagation in each image.

The result of this process is visualized in Fig.~\ref{fig:methods}(e), which clearly shows corresponding regions of scale assignments: the same regions are assigned with high (low) scales in the two images. \\

\noindent{\bf Comparison with~\cite{furukawa2010accurate}:} It is instructional to compare the process described here with the one used for 3D reconstruction from multiple views in~\cite{furukawa2010accurate}. They too begin with feature point extraction and sparse correspondence estimation. Their correspondences are used to build a preliminary 3D point cloud and estimates for the camera matrices of each input image. A continuous 3D surface is then produced by an ``expansion'' process which uses the initial correspondences to seed a search for neighboring matches in an effort to obtain dense correspondences.

We also use an initial, sparse set of correspondences to seed a search for dense correspondences, by propagating information to neighboring pixels. Here, however, we expand the scale estimates, not the correspondences themselves. This is performed for a single pair of images and without going through the process of 3D reconstruction and camera parameter estimation. 

\section{Discussion: Scale accuracy vs. flow accuracy}\label{sec:discussion}
The assumptions underlying our method guarantee that some scales will be repeatable from one image to the next. In particular, the scales at interest points common to both images, in the match-aware propagation of Sec.~\ref{sec:match}, will be covariant and would allow extraction of invariant descriptors. We expect that others, however, may still be inconsistent, resulting in descriptors produced at wrong scales with different feature values. It is therefor reasonable to consider: What effect would wrong scale assignments have on the overall quality of the flow?

To answer this question, we consider the method used for dense correspondence estimation, here, the SIFT-Flow of~\cite{liu2011sift}. It uses belief propagation to minimize the following cost, defined over the estimated flow field (warp) ${\mathbf w}({\mathbf p})=[u({\mathbf p}),v({\mathbf p})]^T$ from each pixel in the source image $I_A$ to its corresponding pixel in the target image $I_B$:

\begin{align}
F({\mathbf w})=&\sum_{\mathbf p}{\min\left(||f(I_A,{\mathbf p}, S_A({\mathbf p}))\right.} \nonumber \\
&~~~~~~~~~~~~\left.{}-f(I_B,{\mathbf p}+{\mathbf w}({\mathbf p}), S_B({\mathbf p}+{\mathbf w}({\mathbf p})))||_1,k\right)\nonumber \\
&+\sum_{\mathbf p}{\nu\left(|u({\mathbf p})|+|v({\mathbf p})|\right)}\label{eq:siftflowcost}\\
&+\hspace{-0.4cm}\sum_{({\mathbf p_1},{\mathbf p_2}\in N)}[~{\min\left(\alpha|u({\mathbf p_1})-u({\mathbf p_2})|,d\right)}+ \nonumber \\
&~~~~~~~~~~~~{\min\left(\alpha|v({\mathbf p_1})-v({\mathbf p_2})|,d\right)}~] \nonumber
\end{align}

\noindent Here, $k$ and $d$ are constant threshold values and $N$ defines a neighboring pixel relationship (e.g., ${\mathbf p_1}$ and ${\mathbf p_2}$ are nearby). The function $f$ represents the SIFT feature transform, where we make explicit the scales used for computing the descriptors, represented by the scale-maps $S_A$ and $S_B$. 

\begin{figure}[!h]
\begin{center}
\includegraphics[width=\linewidth,clip,trim = 0mm 0mm 0mm 0mm]{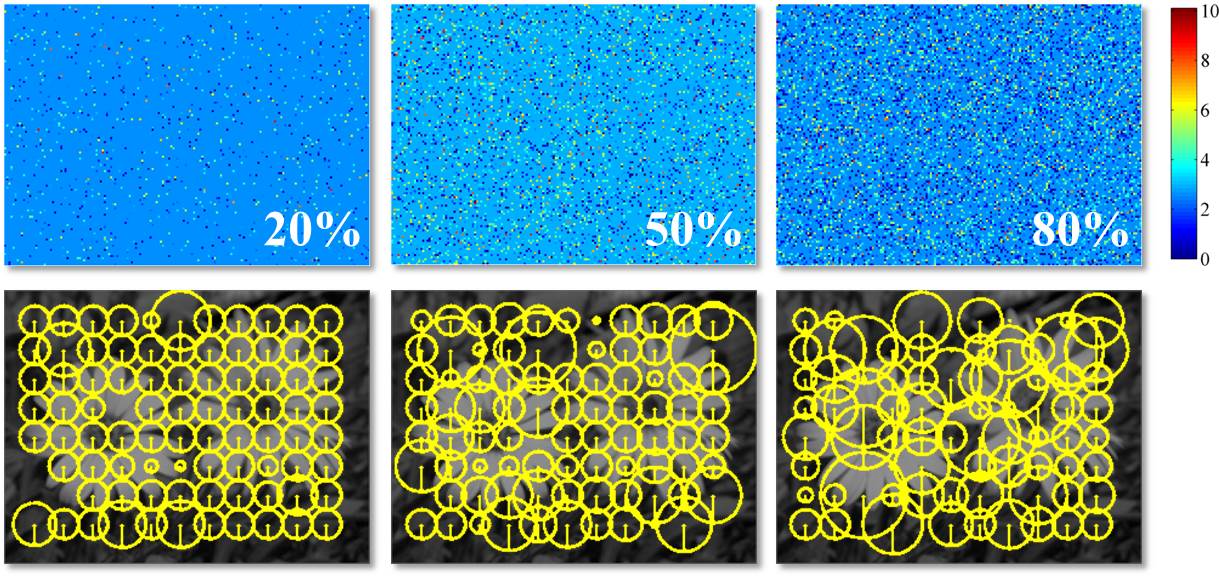}\\%
\includegraphics[width=.95\linewidth,clip,trim = 0mm 0mm 0mm 0mm]{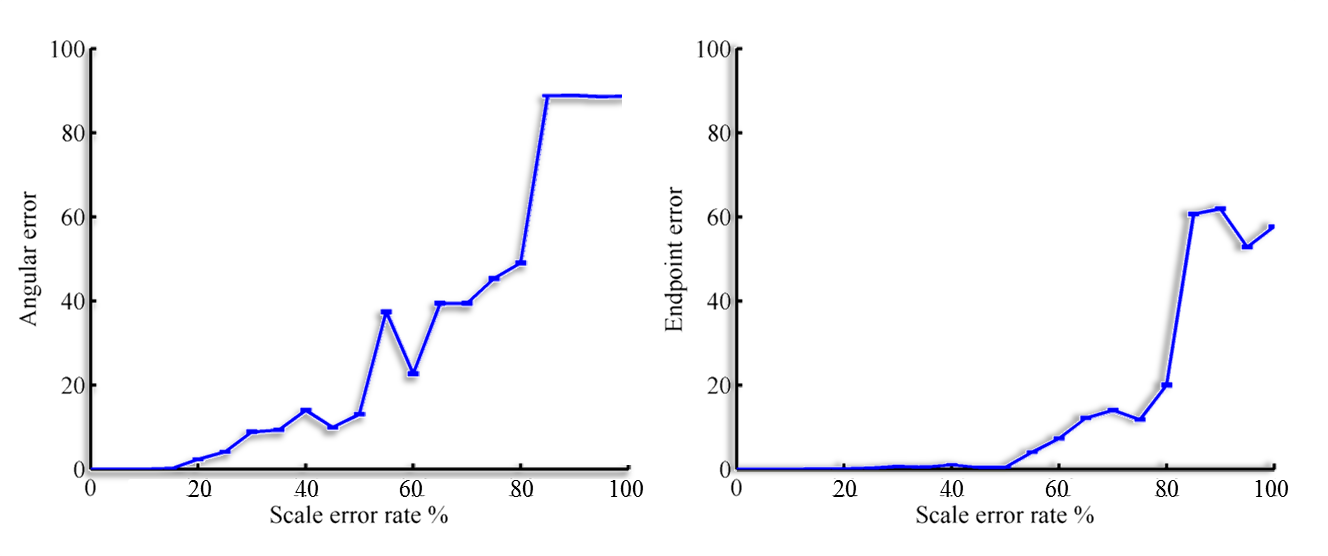}%
\caption{{\bf Effect of wrong scale estimates on flow accuracy using SIFT-Flow~\cite{liu2011sift}.} Top: Scale-maps for 20$\%$, 50$\%$, and 80$\%$ scale assignment errors, visualized by color coding scales (color-bar on the right). The correct scale is the default value of 2.667 for all pixels. Mid: Visualizing the assigned scales, for every 15th pixel. Bottom: Angular errors (left) and endpoint errors (right), $\pm$ SE, for increasing errors in scale estimates. Evidently, flow remains accurate up until about 20$\%$ errors rates.}\label{fig:plot_scale_errors}
\end{center}
\end{figure}

The second term in Eq.~\ref{eq:siftflowcost} represents a requirement for small displacement. The third term, reflects a requirement for a smooth flow-field. Only the first term is affected by scale estimates, and so presumably, the minimization of Eq.~\ref{eq:siftflowcost} should be at least partially robust to scale estimate errors. In practice, the success of SIFT-Flow using Dense-SIFT (DSIFT) descriptors, implies that this is indeed the case: DSIFT uses a single, arbitrarily selected scale for all image pixels, and so one would expect that at least some pixels would have wrong scale estimates. 

We empirically evaluate this tie between scales and accurate flow estimates, in order to obtain a measure of the robustness of SIFT-Flow to scale estimation errors. To this end, we compute the SIFT-Flow between images and themselves using increasing amounts of scale assignment errors. 

Initially, the same constant scale is used for all pixels in each image pair. Using the default parameters of the SIFT extraction routine of~\cite{vedaldi2010vlfeat}, we take the SIFT bin size to be 8 pixels and the magnification factor to be 3, resulting in a scale value of $8/3=2.667$. We then progressively add noise to the scale-map of the target image by randomly selecting increasing numbers of pixels and adding Gaussian noise, with mean zero and STD of 2, to their assigned scales. 

Fig.~\ref{fig:plot_scale_errors}(top) shows scale-maps with noise added to $20\%$, $50\%$, and $80\%$ of the pixels. These synthetically modified scale-maps were used to extract SIFT descriptors (visualized in Fig.~\ref{fig:plot_scale_errors}(mid)), which were then matched using SIFT-Flow. The quality of the resulting flow is evaluated by considering the angular and endpoint errors~\cite{middlebury-flow}. 

Fig.~\ref{fig:plot_scale_errors}(bottom) plots the effect of wrong scale estimates vs. these two errors measures ($\pm$ SE not shown as it was very small). Evidently, the endpoint errors reported in Fig.~\ref{fig:plot_scale_errors}(bottom) remain almost zero, up until a rate of half the image pixels being assigned with wrong scales. Angular errors appear more sensitive to the noise, beginning to grow at 20$\%$ scale assignment errors.

These experiments are synthetic: In a practical scenario, simply resizing one of the images would result in {\em all} its pixels being assigned wrong scales. Fig.~\ref{fig:plot_scale_errors} suggests that in such cases dense correspondence estimation would fail completely, which was indeed shown to be true for SIFT-Flow in~\cite{hassner2012sifts}. The figure also suggests, however, that it may be sufficient to bring scale assignment errors down to only $20\%$ in order for accurate dense correspondences to be obtained.

%

\section{Experiments}\label{sec:results}
We tested our proposed methods on tasks involving images from different scenes (Sec.~\ref{sec:quality}), and stereo with scale changes (Sec.~\ref{sec:middlebury}). These tests demonstrate the application of our method for the tasks of stereo matching and image hallucination\footnote{More results are reported in a longer version of this paper, submitted for publication. Please see author home-page, for an updated version:~\url{www.openu.ac.il/home/hassner}}. 

Our experiments all use SIFT-Flow~\cite{liu2011sift} to compute the dense correspondences, varying the representations used in order to compare the following alternatives: Dense SIFT (DSIFT)~\cite{vedaldi2010vlfeat}; Scale Invariant Descriptors (SID)~\cite{kokkinos2008scale}; Scale-Less SIFTs (SLS)~\cite{hassner2012sifts}; and the two segmentation aware descriptors of~\cite{trulls2013dense}, the segmentation aware SID (Seg. SID) and the segmentation aware SIFT (Seg. SIFT). In all cases, we used the code published by the respective authors of each method, with their recommended parameters unchanged. These methods were compared against our own geometric scale propagation (Geo.), image-aware propagation (Image) and match-aware propagation (Match).

\subsection{Implementation and run-time}\label{sec:runtime}
We implemented all three versions of our scale propagation technique in MATLAB. The multi-scale feature detections used by our proposed methods were obtained using the standard SIFT detector, implemented in the vlfeat library~\cite{vedaldi2010vlfeat}. Minimizing the sparse system of equations resulting from the cost of Eq.~\eqref{eq:cost} was performed using the built-in MATLAB solver, computed on neighborhoods of $3\times 3$ pixels. Finally, scale-varying, dense SIFT descriptors were extracted with vlfeat~\cite{vedaldi2010vlfeat}.


Run-time was measured on an Intel Core i5 CPU, 1.8GHz, with 4GB of RAM and running 64Bit Windows 8.1. We use very small images for these tests ($78\times 52$ pixels) in order to avoid measuring run-time required for swapping memory, when using the more memory intensive representations (SID and SLS).



Descriptor sizes and flow-estimation run-times are summarized in Table~\ref{tab:runtime}. Descriptor dimensions were those measured in practice when running the code provided by the authors of each method. We extract a single, 128D SIFT descriptor per pixel -- the same storage required by the standard SIFT-Flow method, and {\em an order of a magnitude} less storage than required by both the SID and SLS representations. Not surprisingly, the time required for establishing flow using our own method is the same as the time required for the original SIFT-Flow, an order of magnitude less than the SID descriptors, and two orders of magnitude less than SLS. 

\begin{table}[!h]
\begin{center}
\caption{{\bf Comparison of different descriptor dimensions, and flow-estimation run-time.} Mean run-times were measured using SIFT-Flow, on $78\times 52$ pixel images.} \label{tab:runtime}
\begin{footnotesize}
\begin{tabular}{lcc}
\toprule
Method & Flow run-time (sec.) & Dim. \\
\hline
DSIFT~\cite{liu2011sift} & 0.8 & 128D  \\
SID~\cite{kokkinos2008scale} & 5 & 3,328D \\%
SLS~\cite{hassner2012sifts} & 13 & 8,256D \\
Seg. SID~\cite{trulls2013dense}  & 5 & 3,328D \\
\hline
Us & 0.8 & 128D \\
\bottomrule
\end{tabular}
\end{footnotesize}
\end{center}
\end{table}

Finally, we compared the time required for optimizing our cost function of Eq.~\eqref{eq:cost} (propagating the scales) with the time required by SIFT-Flow to estimate correspondences. Here, we varied the size of the images from the original $78\times 52$ pixels to $780\times 520$ pixels. For all image sizes, scale propagation required less than 7$\%$ of the time for computing the correspondences themselves, using SIFT-Flow. Consequently, SIFT-Flow performed following scale propagation requires only slightly more time than runing SIFT-Flow once, without scale propagation.

\subsection{Qualitative results}\label{sec:quality}
\begin{figure*}[!h]
\begin{center}
\includegraphics[width=\linewidth,clip,trim = 0mm 0mm 0mm 0mm]{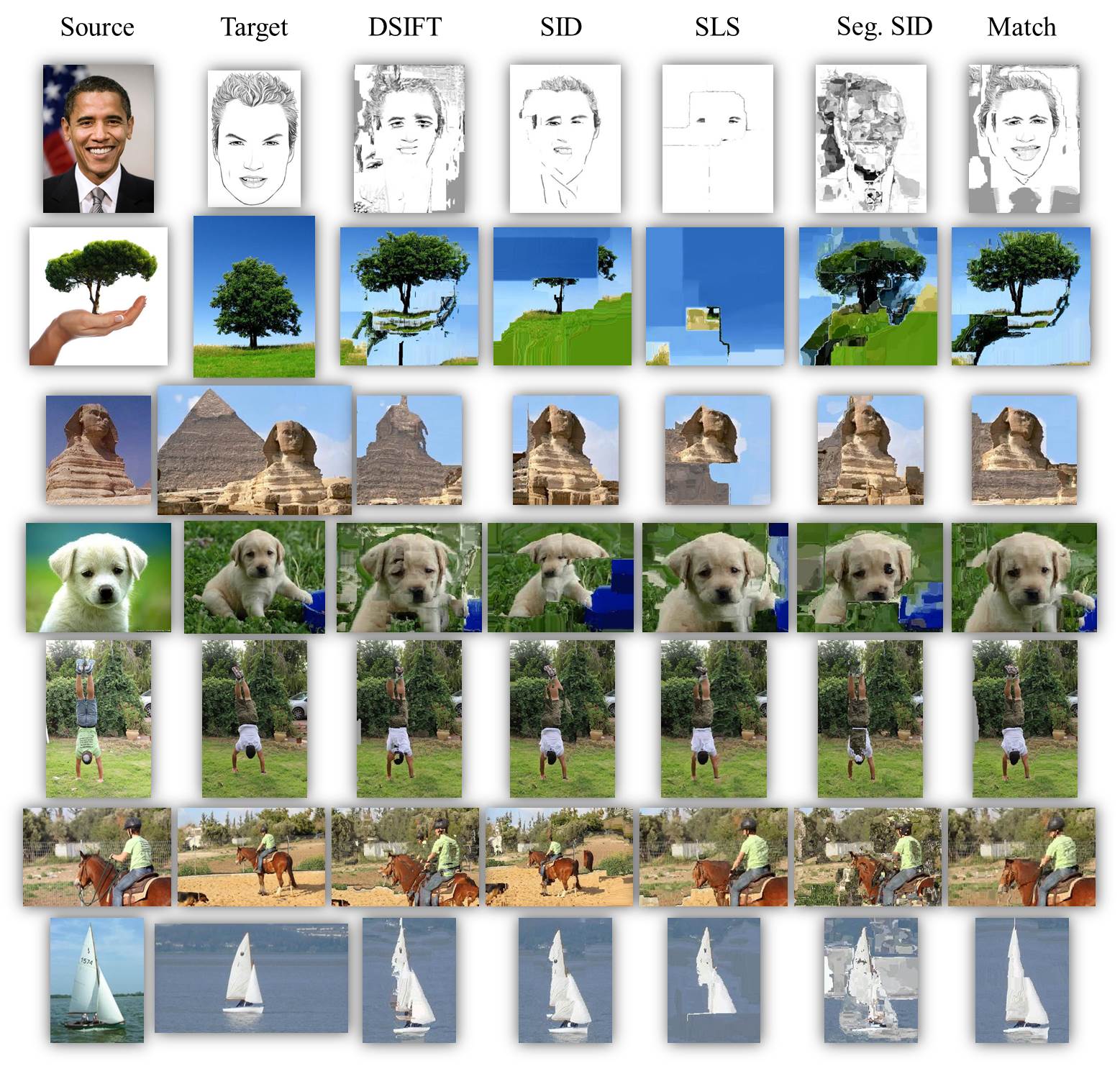}\\%
\caption{{\bf Image hallucination results.} Each row presents dense correspondences established from source images to target images, illustrated by warping the target photos back to the sources using the estimated flows. We compare the following representations, from left to right: DSIFT~\cite{vedaldi2010vlfeat}, SID~\cite{kokkinos2008scale}, SLS~\cite{hassner2012sifts}, Segmentation aware SID (Seg. SID)~\cite{trulls2013dense}, and SIFT descriptors extracted using our own Match-aware scale propagation. Good results should have the colors of the target photos, warped to the shapes appearing in the source photos.}\label{fig:qualitative}
\end{center}
\end{figure*}

\begin{figure*}[!h]
\begin{center}
\includegraphics[width=\linewidth,clip,trim = 0mm 0mm 0mm 0mm]{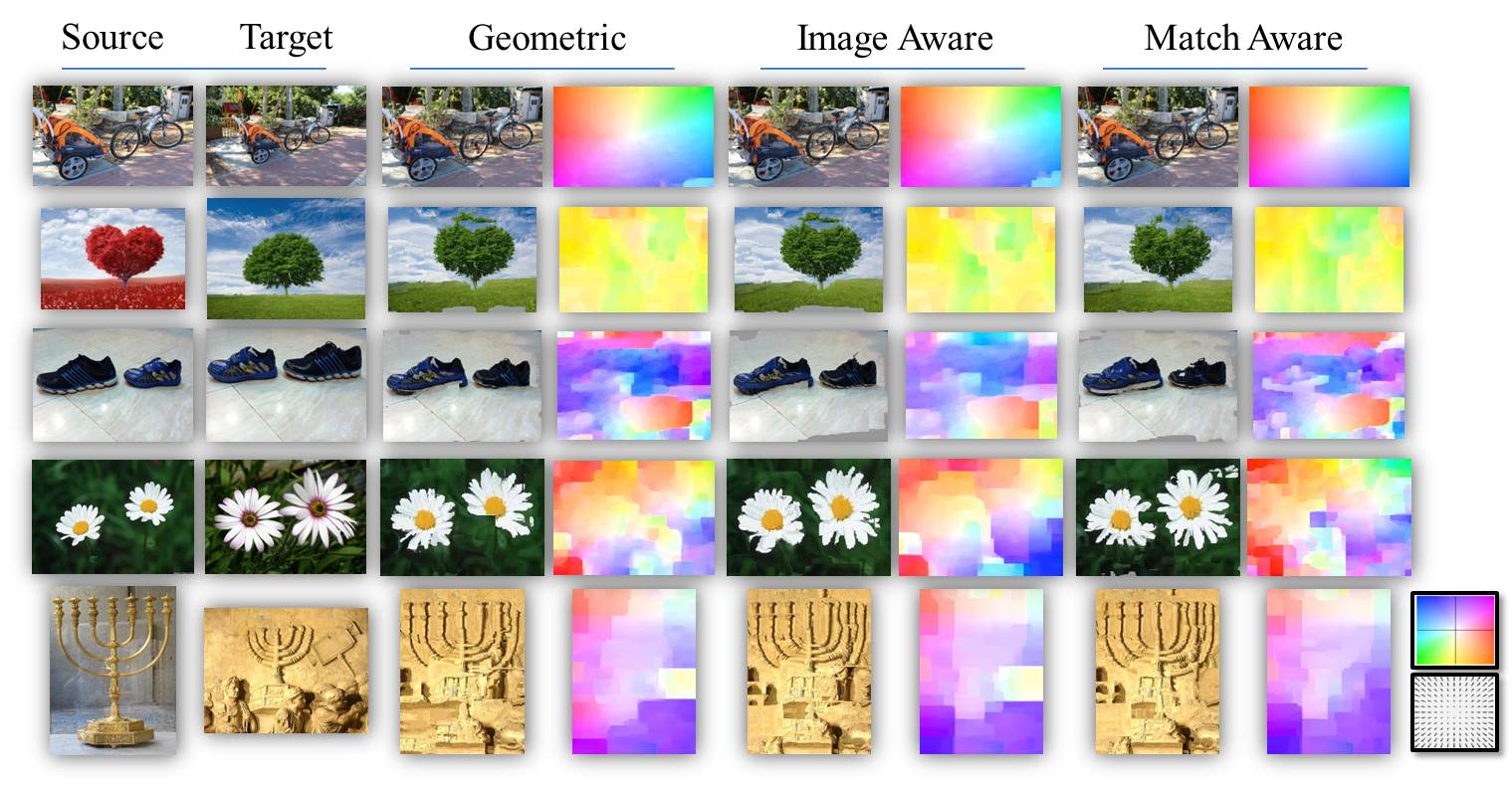}\\%
\caption{{\bf Image hallucination results - comparison of proposed methods.} Each row presents dense correspondences established from source images to target images, illustrated by warping the target photos back to the sources using the estimated flows. We compare our three proposed methods for propagating scales, from left to right: Geometric scale propagation (Sec.~\ref{sec:geometric}), image-aware propagation (Sec.~\ref{sec:image}), and match-aware propagation (Sec.~\ref{sec:match}). Each hallucination result provides also a visualization of the estimated flow field. Flow legend is provided on the bottom right.}\label{fig:qualitative_compare}
\end{center}
\end{figure*}

Figures~\ref{fig:teaser},~\ref{fig:qualitative} and~\ref{fig:qualitative_compare} present image hallucination results obtained by computing dense correspondences from source to target images, and then warping the target colors back to the sources using the estimated flows. In all cases, good results would have the target image colors warped to the shapes appearing in the source photos.

The results included in these figures were all selected in an effort to reflect the most challenging instances of the dense correspondence estimation problem. Image pairs exhibit extreme variations in local scales, different scenes, different viewing conditions and more. We additionally emphasize cases where images have large homogenous regions. Existing feature detectors typically cannot estimate local scales in such image regions. By propagating scale estimates, we allow for scale-invariant descriptors to be extracted and dense correspondences to be estimated even in such cases. 

Fig.~\ref{fig:qualitative_compare} provides a comparison of the three proposed methods of propagating scales: Geometric scale propagation (Sec.~\ref{sec:geometric}), image-aware propagation (Sec.~\ref{sec:image}), and match-aware propagation (Sec.~\ref{sec:match}). Evidently match-aware propagation provides the most coherent results, though its two simpler alternatives are comparable in the quality of their results. 

Though the results obtained with our Match-aware scale propagation (Fig.~\ref{fig:qualitative}, rightmost column) are sometimes qualitatively similar to those obtained by other representations, ours consistently produces good results. This, despite much lower run-time and storage requirements compared to the scale-invariant descriptors, SID, SLS, and Seg. SID. Unsurprisingly, DSIFT performs worst when applied to image pairs with scale changes.

\begin{table*}[t]
\begin{center}
\caption{{\bf Results on the scaled-Middlebury benchmark}. Angular errors (AE) and endpoint errors (EE), $\pm$ SD, on resized images from the Middlebury benchmark~\cite{middlebury-flow}. Lower scores are better; shaded cells are best scoring. Please see text for more details.} \label{tab:middlebury}
\begin{footnotesize}
\begin{tabular}{lccccccccc}
\toprule
Data &DSIFT~\cite{liu2011sift}&SID~\cite{kokkinos2008scale}&SLS~\cite{hassner2012sifts}&Seg. SIFT~\cite{trulls2013dense} & Seg. SID~\cite{trulls2013dense} &Geo.&Image&Match\\
\hline
&\multicolumn{8}{c}{Angular Errors $\pm$ SD}\\
\hline
Dimetrodon	&3.13 $\pm$ 4.0 	&0.16  $\pm$0.3 &0.17  $\pm$0.5 				&2.45 $\pm$ 2.8 	&0.23 $\pm$ 0.7 & 0.61 $\pm$ 0.7 &2.95 $\pm$ 4.2 &\cellem  0.15 $\pm$ 0.3 \\
Grove2			&3.89 $\pm$ 11.9 	&0.66  $\pm$4.4 &0.15  $\pm$0.3 				&4.77 $\pm$ 15.3 	&0.22 $\pm$ 0.6 & 2.30 $\pm$ 2.3 &1.78 $\pm$ 2.1 &\cellem  0.13 $\pm$ 0.3 \\
Grove3			&2.67 $\pm$ 2.8 	&1.62  $\pm$6.9 &\cellem 0.15 $\pm$0.4	&8.93 $\pm$ 15.6 	&0.22 $\pm$ 0.6 & 6.26 $\pm$ 19.3 &1.72 $\pm$ 2.1 &0.18 $\pm$ 0.4 \\
Hydrangea		&9.76 $\pm$ 18.0 	&0.32  $\pm$0.6 &0.22  $\pm$0.8 				&7.10 $\pm$ 10.6 	&0.23 $\pm$ 0.7 & 1.72 $\pm$ 2.3 &6.25 $\pm$ 11.6&\cellem  0.19 $\pm$ 0.4 \\
RubberWhale	&5.27 $\pm$ 8.6 	&0.16  $\pm$0.3 &0.15  $\pm$0.3 				&6.13 $\pm$ 17.2 	&0.16 $\pm$ 0.3 & 1.56 $\pm$ 2.1 &3.31 $\pm$ 5.4 &\cellem  0.13 $\pm$ 0.2 \\
Urban2			&3.65 $\pm$ 10.7 	&0.37  $\pm$2.7 &0.32  $\pm$1.3 				&2.82 $\pm$ 4.1 	&0.25 $\pm$ 1.1 & 0.53 $\pm$ 0.8 &4.28 $\pm$ 6.8 &\cellem  0.21 $\pm$ 0.5 \\
Urban3			&3.87 $\pm$ 5.1 	&0.27  $\pm$0.6 &0.35  $\pm$0.9 				&3.53 $\pm$ 4.4 	&0.31 $\pm$ 1.0 & 1.43 $\pm$ 1.96 &3.79 $\pm$ 7.9 &\cellem  0.24 $\pm$ 0.5 \\
Venus				&2.66 $\pm$ 2.9 	&0.24  $\pm$0.6 &\cellem  0.23 $\pm$0.5	&2.77 $\pm$ 6.7 	&\cellem  0.23 $\pm$ 0.5 &1.32 $\pm$ 1.2 &2.43 $\pm$ 2.3 &0.30 $\pm$ 0.6 \\
\toprule
&\multicolumn{8}{c}{Endpoint Errors  $\pm$ SD}\\
\hline
Dimetrodon	&10.97 $\pm$ 8.7 	&\cellem 0.7 $\pm$0.3 &0.8  $\pm$0.4 &10.34 $\pm$ 7.5 &0.97 $\pm$ 1.1 &2.72 $\pm$ 1.5 &11.21 $\pm$ 10.2 &0.81 $\pm$ 0.3 \\
Grove2			&14.38 $\pm$ 11.5 &1.5 $\pm$5.0 	&0.77  $\pm$0.4 &15.50 $\pm$ 11.0 &1.05 $\pm$ 1.9 &12.8 $\pm$ 10.2 &9.06 $\pm$ 9.4 &\cellem  0.71 $\pm$ 0.3 \\
Grove3			&13.83 $\pm$ 9.7 	&4.48  $\pm$10.5 &\cellem  0.87  $\pm$0.4 &24.33 $\pm$ 20.0 &1.37 $\pm$ 3.3 &14.4 $\pm$ 14.7 &9.22 $\pm$ 7.7 &1.15 $\pm$ 2.5 \\
Hydrangea		&25.32 $\pm$ 17.1 &1.59  $\pm$2.8 &0.91  $\pm$1.1 &24.21 $\pm$ 17.3 &0.88 $\pm$ 0.6 &10.2 $\pm$ 8.9 &15.69 $\pm$ 19.2 &\cellem  0.79 $\pm$ 0.5 \\
RubberWhale	&22.59 $\pm$ 15.8 &0.73  $\pm$1.1 &0.8  $\pm$0.4 &17.33 $\pm$ 14.8 & 0.73 $\pm$ 0.4 &7.63 $\pm$ 8.5 &11.27 $\pm$ 15.6 &\cellem  0.68 $\pm$ 0.3 \\
Urban2			&18.96 $\pm$ 17.5 &1.33  $\pm$3.8 &1.51  $\pm$5.4 &13.36 $\pm$ 10.3 & 1.21 $\pm$ 3.7 &2.73 $\pm$ 1.7 &15.51 $\pm$ 15.2 &\cellem  0.96 $\pm$ 1.4 \\
Urban3			&19.83 $\pm$ 17.1 &1.55  $\pm$3.7 &9.41  $\pm$24.6 &15.44 $\pm$ 11.5 & 1.47 $\pm$ 4.1 &6.10 $\pm$ 4.9 &14.91 $\pm$ 15.0 &\cellem  1.15 $\pm$ 1.8 \\
Venus				&9.86 $\pm$ 8.7 	&1.16  $\pm$3.8 &\cellem  0.74  $\pm$0.3 &11.86 $\pm$ 11.4 &\cellem  0.74 $\pm$ 0.5 &4.25 $\pm$ 2.0 &10.92 $\pm$ 11.5 &0.82 $\pm$ 0.4 \\
\bottomrule
\end{tabular}
\end{footnotesize}
\end{center}
\end{table*}

\subsection{Scaled-Middlebury results}\label{sec:middlebury}
We repeat the qualitative experiments reported in~\cite{hassner2012sifts}, measuring the accuracy of stereo correspondences in the presence of extreme scale changes. We use the well-known Middlebury data set~\cite{middlebury-flow}, containing pairs of images of the same scenes, acquired from different viewpoints. Since these images do not include scale changes these are introduced by re-sizing both images in each pair, one to 0.7 its size and one to 0.2 (the original sizes are not used, due to limitations of memory for the SLS and SID descriptors). Our tests include the image pairs with ground truth dense correspondences, which we use to compute Angular Error (AE) and Endpoint Error (EE) rates, along with standard deviations ($\pm$ SD)~\cite{middlebury-flow} for each of the representations tested.

Our results are reported in Table~\ref{tab:middlebury}. These demonstrate that by propagating scales we achieve better accuracy on almost all of the tested pairs, falling in only slightly behind the far more expensive multi-scale representations, when this is not the case.

\section{Conclusions}\label{sec:conc}
Modern computer vision systems owe much of their success to the development of effective scale selection techniques, key to the extraction of local, scale-invariant descriptors. These widely used techniques have focused almost entirely on the few image locations where local appearance variations provide sufficient cues for selecting reliable (repeatable) scales. In contrast, we propose a means for determining reliable scales for {\em all} the pixels in the image, regardless of their local appearances. 

We describe three means of propagating scales from pixels selected by a standard, multi-scale, feature detector to all other image pixels. Our approach allows for truly scale-invariant dense SIFT descriptors to be extracted and then matched between images. An important aspect of our method, is that unlike alternatives proposed in the recent past, it makes very little computation and storage requirements beyond those needed for matching standard, non scale-invariant, dense SIFT descriptors. The result is a practical, effective, and efficient method for establishing dense correspondences across scenes, which does not make any assumptions on local scale variations in the images being matched.

Our method was tested qualitatively, by producing image hallucination results for challenging image pairs, as well as quantitatively for its flow accuracy. These have all shown how propagating scales contributes to reliable and robust dense correspondence estimation. \\

\noindent{\bf Future work.} This paper opens a number of prospective directions for future research. One immediate direction is to explore how well other transformations, chiefly local orientation, may benefit from a similar approach. Our initial experiments conducted by adding an orientation-map, analogous to the scale-maps used here, were inconclusive. We believe this is because rotation may be a more global phenomenon compared to scale; rotations are often applied to entire images whereas scales frequently change from one portion of the image to another. Further study is required to see if and how orientation can also benefit from a similar approach.

Applications of dense correspondence estimation were surveyed in Sec.~\ref{sec:prev}. An additional line of work would be to explore the impact of our method's improved robustness on existing and new label transfer tasks. 

Finally, showing that robust dense correspondences can be established with reasonable computation and storage requirements, raises intriguing questions regarding the possible roles of dense correspondence estimation in biological vision. Correspondence estimation is well known to play a key part in depth perception by stereo vision. The success of label transfer approaches in computer vision suggests that it may be worth while to explore the existence of similar mechanisms in biological visual systems.

\ifCLASSOPTIONcaptionsoff
  \newpage
\fi

\footnotesize{
\bibliographystyle{IEEEtran}
\bibliography{scalemaps_a}
}
\end{document}